\documentclass[11pt]{article}
\usepackage[margin=1in]{geometry}
\usepackage[T1]{fontenc}
\usepackage{lmodern,microtype}
\usepackage{amsmath,amssymb,amsthm,mathtools,bm}
\usepackage{booktabs,array,tabularx}
\usepackage{needspace,etoolbox,comment}
\usepackage{xurl}
\usepackage[ruled,vlined,linesnumbered]{algorithm2e}
\usepackage[authoryear,round]{natbib}
\usepackage{xcolor}
\usepackage[colorlinks=true,linkcolor=blue!50!black,citecolor=blue!50!black,urlcolor=blue!50!black]{hyperref}
\hypersetup{pdftitle={Detection boundaries and adaptive inference for textGrain},pdfauthor={}}
\newtheorem{theorem}{Theorem}[section]
\newtheorem{proposition}[theorem]{Proposition}

\newcommand{\tg}{\texttt{textGrain}}
\newcommand{\nbsb}[1]{\textcolor{red}{\texttt{SB:[#1]}}}
\newtheorem{corollary}[theorem]{Corollary}
\theoremstyle{definition}

\newcommand{\email}[1]{\texttt{#1}}
\newtheorem{definition}[theorem]{Definition}
\newtheorem{assumption}{Assumption}
\newtheorem{example}{Example}

\theoremstyle{remark}
\newtheorem{remark}[theorem]{Remark}
\DeclareMathOperator{\KL}{D_{\mathrm{KL}}}
\DeclareMathOperator{\Var}{Var}

\DeclareMathOperator{\TV}{TV}

\newcommand{\E}{\mathbb E}
\newcommand{\Pp}{\mathbb P}
\newcommand{\bz}{\mathbf Z}
\newcommand{\IP}{\mathbb{P}} 
\newcommand{\ind}{\mathbf 1}
\newcommand{\wall}{\mid\mid}
\newcommand{\Unif}{\operatorname{Unif}}

\newcommand{\cW}{\mathcal W}
\newcommand{\cF}{\mathcal F}
\def\bo{\mathbf{1}}
\newcommand{\iidsim}{\overset{\text{i.i.d}}{\sim}}
\newcommand{\cH}{\mathcal H}

\newcommand{\dd}{\,\mathrm d}
\newcommand{\blk}{\mathrm{blk}}
\newtheorem{infthm}[theorem]{Informal Theorem}
\definecolor{tgNavy}{HTML}{1D3557}     
\definecolor{tgGold}{HTML}{E39B00}     
\definecolor{tgRed}{HTML}{C0392B}      
\definecolor{tgYellow}{HTML}{F7C873}   
\definecolor{tgGreen}{HTML}{8FD3B6}    
\definecolor{tgGrey}{HTML}{C4C4C4}     
\definecolor{tgBlue}{HTML}{2A3B78}     
\definecolor{tgLavender}{HTML}{B7A3E6} 
\definecolor{tgPurple}{HTML}{2E1A5C}   
\newcommand{\cw}[2]{\textcolor{#1}{\textbf{#2}}}
\newcommand{\R}{\mathbb{R}}      
\SetKwInput{KwInput}{Input}
\SetKwInput{KwOutput}{Output}
\SetKw{Return}{return}
\allowdisplaybreaks[1]
\title{\tg\ watermarking is harder to detect: can sum-tests still do it optimally?}
\author{
    Swapnaneel Bhattacharyya \hspace{1.5em} Soham Bonnerjee
}
\date{
    \normalsize  University of Pennsylvania\\[0.3em]
    \small\email{\{swap, sohamb\}@wharton.upenn.edu}
}

\begin{document}
\maketitle
    \begin{abstract}
Watermarking embeds a signal in text generated by large language models that is imperceptible to readers but statistically detectable with a secret key. The recently proposed \tg\ watermark calibrates this signal against the model's sampling entropy through an explicit entropy budget, combining the stochasticity of Red-Green watermarks with the unbiasedness of Gumbel-max watermarks. The accompanying report \cite{textgrain2026} proposes a detector that sums pivotal statistics. We study when, and how efficiently, \tg\ can be detected after human edits, which we model as a sparse mixture in which a fraction $\varepsilon_n\asymp n^{-p}$ of positions retain the watermark and token uncertainty decays as $n^{-q}$. First, we show that entropy calibration comes at a cost: the minimax detection boundary of \tg\ is $p+q=1/2$, so its detectable region is strictly smaller than that of Gumbel-max and maximal-coupling watermarks, whose boundary is $2p+q=1$. This cost stems from the construction of \tg\ via finitely many columns rather than from entropy calibration itself: we exhibit an unbiased scheme with the same entropy loss that remains detectable where \tg\ is not. Second, and perhaps surprisingly, this loss makes the simple sum detector optimal at the boundary when the number of transport columns is fixed, unlike for Gumbel-max. When the number of columns grows and the entropy budget does not vanish, detection becomes easier, and the detectable region expands while that of the sum detector does not. Third, for a fixed edit fraction, we characterize the optimal exponential rate at which the type II error can decay, and show that the sum detector falls strictly short of it at every fixed alternative. Finally, we propose a weighted multiscale binomial scan that is valid in finite samples and requires no knowledge of the next-token distributions, the entropy budget, or the edit fraction; it attains every detection boundary above, as well as the optimal exponent.
\end{abstract}

\section{Introduction} \label{se:intro}
To preserve content authenticity and curb misuse of AI, an LLM-generated text is often watermarked, which imbues it with a signal that is statistically verifiable, thus aiding detection, but otherwise remains imperceptible. \tg\ \citep{textgrain2026} represents the latest in a series of developments \citep{kirchenbauer2023watermark, aaronson2023watermarking, christ2024undetectable, kuditipudi2024robust, hu2024unbiased, wu2024dipmark, dathathri2024scalable, zhao2025permute} to enhance the quality of such watermarked texts while allowing for statistical detection via a suitable test statistic. Introduced in compliance with the EU content provenance rules \citep{euaiact2024, eucode2026}, \tg\ marries the stochasticity of the Red-Green watermark \citep{kirchenbauer2023watermark} with the unbiasedness of the Gumbel-max watermark \citep{aaronson2023watermarking} to conjure watermarks that respect the NTP inherent to the large language model, while reducing repetitive generation. Thematically, the \tg\ mechanism speaks to several recent threads in the literature. For example, the tendency of the Gumbel-max watermark to yield identical outputs for identical prompts has motivated diversity-preserving variants \citep{fu2024gumbelsoft}; the observation that low-entropy tokens carry little watermark signal has led to entropy-thresholded watermarking \citep{lee2024wrote} and entropy-weighted detection \citep{lu2024entropy}. The apparent trade-off between detection ability and model distortion has been analyzed as a constrained optimization problem by \citet{cai2026towards}. \tg\ mollifies each of these concerns by calibrating the trade-off between watermark signal and sampling entropy through an explicit entropy budget. The technical report \citet{textgrain2026} builds on prior work on optimal-transport based watermarks for low entropy distributions \cite{tsur2026heavywater}, as well as debiasing Red-Green watermarking via maximal coupling \citep{xie2025debiasing}, and discusses the detailed mechanism behind \tg. Along the way, it also touches upon a detection strategy based on sums of pivot statistics; however, the optimality of such detection strategies under human edits is yet unaddressed. We complement \citet{textgrain2026} on this front by providing a characterization of optimal detection for \tg-generated texts. Our work builds on recent research on optimal watermark detection \citep{li2025framework, li2026robust, cai2025optimal, blanchet2026optimal}, and unearths the following simple insights:
\begin{itemize}
    \item \tg\ \textit{makes watermark detection harder}.
    \item The sum test, proposed as \tg\ detector in \cite{textgrain2026}, is optimal if number of columns in the cost matrix is fixed.
    \item The sum test is \textit{not} optimal if number of columns in the cost matrix grow.
\end{itemize}

Before proceeding further, it is prudent to briefly introduce the set-up and mechanism of \tg. Let $\cW$ be a finite vocabulary with $V=|\cW|\ge2$. For a probability
vector $P=(P_w)_{w\in\cW}$, write $H(P)=-\sum_wP_w\log P_w$ and
$\Delta(P)=1-\max_wP_w$. Fix some $B \ge 2$. At each watermarked position, let $g:\cW\to[B]$ is a partition, such that for each $w \in \cW$, $g(w)$ belongs to one of the $B$ blocks. Further denote the block $b$ probability as
$P_b^{\blk}=\sum_{w:g(w)=b}P_w$, where we unburden the notation by suppressing the dependence on $g$ when it is clear from the context. For some $m \ge 1,$ suppose
$$\Pi(P^{\blk},m^{-1}\bo):= \{ A \in \R^{b\times m}: A \bo = (P_b^\blk)_{b\in [B]}, A^\top \bo = m^{-1}\bo\},$$ 
 be the set of couplings over $[B] \times [m]$ with marginals $P^{\blk}$ and $\Unif([m])$. Here $\bo=(1,\ldots, 1)^\top$. Consider $\bz=(Z_{bj})_{b\in [B], j \in [m]}\iidsim \Xi:=$ standard Gumbel variables. These are also called pseudo-random variables, since, given tokens $w_1\ldots w_{t-1}$ and the NTP $P_t$, both the partition $g$ and the Gumbel random variables can be deterministically constructed via a $\mathsf{Key}$ and a hash function $\mathcal{A}$. Nevertheless, we invoke the cryptographic properties of $\mathcal{A}$ \citep{barak2021intensive, schneier1996applied, wu2024dipmark, zhao2025permute} to treat $\bz$ as random. The deterministic construction comes handy since a detector employs $\mathsf{Key}$ and $\mathcal{A}$ to make $g$ and $(Z_{bj})_{b\in [B], j \in [m]}$ available to it. 

\noindent
Write $G_{bj}=(Z_{bj}-\gamma)/(\pi/\sqrt6)$, where
$\gamma=\E Z_{11}$ is the mean of a standard Gumbel variable.
For $\beta\in[0,1]$, consider the entropic OT coupling
\begin{equation}
\label{eq:transport}
\pi_{P,\beta}\in
\operatorname*{arg\,max}_{\substack{\pi\in\Pi(P^{\blk},m^{-1} \bo)\\
\KL(\pi\|P^{\blk}\otimes m^{-1} \bo)\le\beta H(P)}}
\sum_{b=1}^B\sum_{j=1}^m\pi(b,j)G_{bj}.
\end{equation}
Then, given $J \sim \Unif([m])$, \tg\ samples a token $W$ according to
\begin{equation}
Q(w\mid J=j,g,G)
=m\pi_{P,\beta}(g(w),j)\frac{P_w}{P_{g(w)}^{\blk}}, \nonumber
\  P_w>0.
\end{equation}
  As for detection, \cite{textgrain2026} proposes the test statistic 
  \begin{equation}
      \label{eq:score_convention}
      T^{\tg}_n:=- \sum_{t=1}^n \log(1- Y_t), \ \text{where, } Y_t:= F_G(Z_{t, g(W_t), J_t}).
  \end{equation}
 This leads to the following natural questions:
  \begin{itemize}
      \item Does the entropy-calibration of \tg\ make watermark detection harder compared to a Gumbel-max or the maximal coupling based debiased Red-Green watermarking schemes\citep{xie2025debiasing} ? In particular, the Gumbel-max lets the keyed randomness completely determine the next token, whereas \tg\ preserves additional sampling randomness even after the key is fixed by allowing for $\beta>0$ in \eqref{eq:transport}. Since watermark detection looks for dependence between tokens and keyed randomness, limiting the key’s influence may weaken statistical evidence, thereby making detection harder.
      \item The second question is more straightforward: is $T^\tg_n$ able to detect \tg-watermarked text whenever such a detection is possible?
      \item Finally, detecting a watermark eventually is not the same as detecting it efficiently. Following \citet{li2026robust}, we ask how fast the type II error of a level-$\alpha$ detector can decay when a fixed fraction of the text remains watermarked, and whether $T_n^{\tg}$ attains this optimal exponential rate. Where it does not, either at the detection boundary or in efficiency, can a single detector that requires no knowledge of the NTPs, the entropy budget, or the edit fraction attain both optimality criteria?
  \end{itemize}
To properly address these questions, we quickly get some notational niceties out of the way. For a fixed $P,\beta,g$, let $f_{1,P,\beta,g}$ denote
the density of $Y$, and define the corresponding time-indexed version
$f_{1,t}=\E[f_{1,P_t,\beta_t,g_t}\mid\cF_{t-1}]$. Following \cite{li2026robust, li2025framework}, we answer the aforementioned questions by formulating the watermark detection as a hypothesis testing problem:
\begin{equation}
\label{eq:mixture}
\begin{split}
H_0 &: Y_t\mid\cF_{t-1}\sim\Unif(0,1), \hspace{.3cm} 1 \leq t \leq n,\\
H_1^{\tg} &: Y_t\mid\cF_{t-1}\text{ has density }
(1-\varepsilon_n)+\varepsilon_n f_{1,t},\  1\le t\le n.
\end{split}
\end{equation}
Human edits, paraphrasing, or mixing with human-written text replace an
unknown fraction of watermarked tokens with tokens that are independent
of the key. Each scored position is therefore watermarked with
probability $\varepsilon_n$ and otherwise carries a uniform pivot, which
yields the mixture density in $H_1^{\tg}$. Here $\varepsilon_n$ measures surviving
text--key dependence, and $\varepsilon_n\to0$ models heavily edited text.
This is known as the \textit{sparse watermark regime}
\citep{li2025optimal, xie2025debiasing, li2026robust}, and is motivated
strongly by the sparse detection problem \citep{donoho2004, donoho2015}.
More concretely, we consider sparse watermark signal
$\varepsilon_n \asymp n^{-p}$ and
$\min_{t\in [n]} \Delta (P_t) \asymp \max_{t\in [n]} \Delta (P_t) \asymp n^{-q}$
with $\Delta(P):= 1- \max_w P_w$. In this setting, we have the following sharp characterization for the detection of \tg watermarking.
 \begin{infthm}\label{inf-thm-boundary}
     \begin{itemize}
         \item[(i)] \textit{Without sufficient watermark signals, i.e. when $p + q>1/2$, no test can asymptotically separate $H_0$ and $H_1^{\tg}$ with non-trivial power.}
         \item[(ii)] \textit{When $p+q<1/2$, i.e. watermark signal is sufficient, $H_0$ and $H_1^{\tg}$ separate asymptotically. Moreover, $H_1^{\tg}$ can be detected using $T_n^\tg$.}
     \end{itemize}
 \end{infthm}
Theorem~\ref{inf-thm-boundary}, stated for a fixed number of columns and an entropy fraction bounded away from zero, already answers our first two questions in this basic setting. Since both Gumbel-max \citep[Theorem 2]{li2026robust} and maximal-coupling watermarks \citep[Theorem 3.5]{xie2025debiasing} are detectable throughout $2p+q<1$, the detectable region of \tg\ is strictly smaller whenever $q>0$, so \tg\ is indeed harder to detect. This shrinkage has a perhaps unexpected consequence: the \tg\ boundary $p+q=1/2$ coincides with the boundary of sum tests for those schemes \citep{li2026robust,xie2025debiasing}, which is why the report's $T_n^{\tg}$ suffices here. Our contributions below examine how far these conclusions extend beyond this setting, and address the third question.

\begin{figure}[!tbp]
\centering
\begin{minipage}[t]{0.48\textwidth}
\vspace{0pt}
\centering
\includegraphics[width=\linewidth]{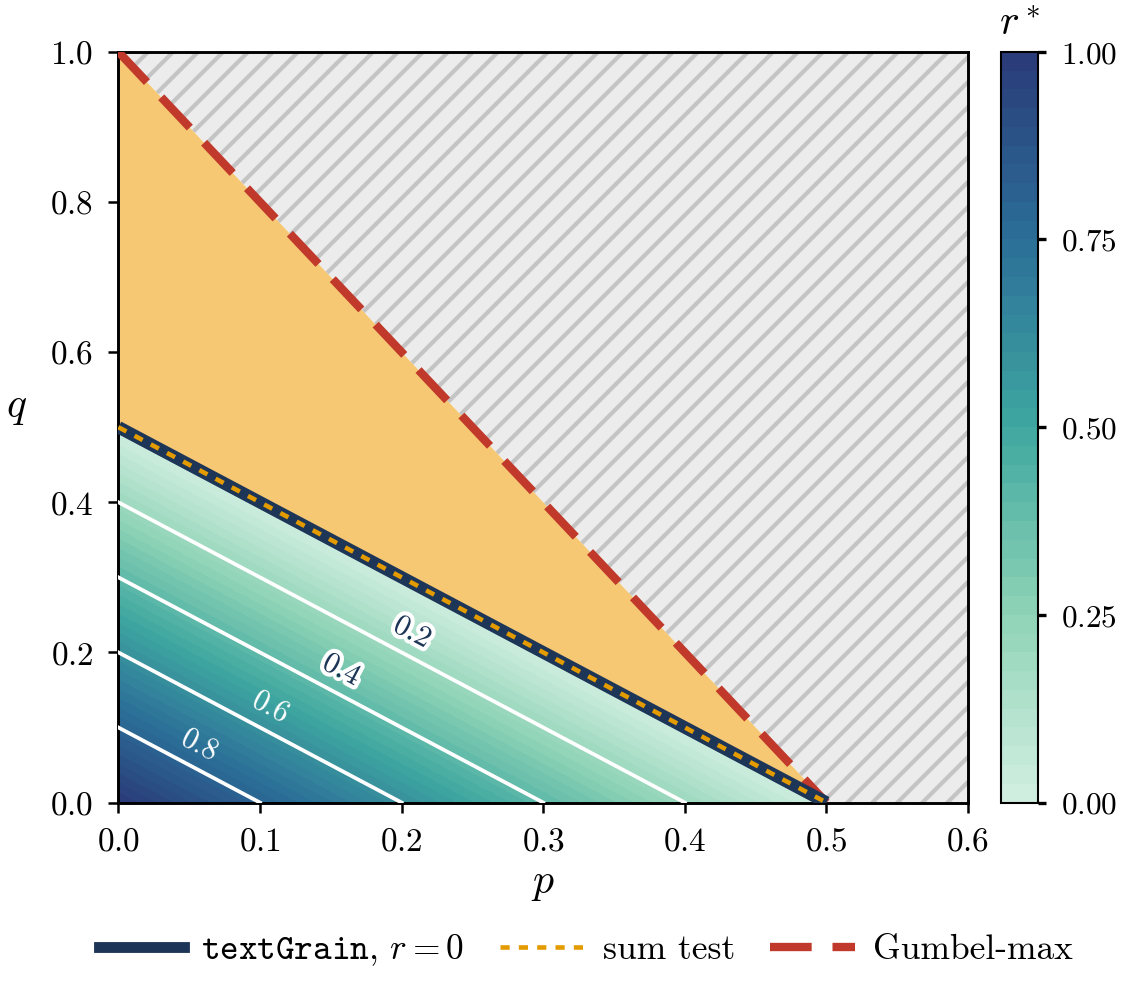}
\caption{\small Phase diagram for \tg\ with a fixed number of columns $m$, where $\varepsilon_n\asymp n^{-p}$ and $\Delta_n\asymp n^{-q}$. The \cw{tgNavy}{navy} line $p+q=1/2$ is the minimax boundary of \tg\ for $\beta_n\asymp1$, that is, $r=0$ (Corollary~\ref{cor:phase}); it coincides with the boundary of $T_n^{\tg}$ (\cw{tgGold}{gold}, dotted; Theorem~\ref{thm:sums}). The \cw{tgRed}{red} dashed line $2p+q=1$ is the boundary of Gumbel-max \citep{li2026robust} and maximal coupling \citep{xie2025debiasing}. Below the \cw{tgNavy}{navy} line, the shading from \cw{tgGreen}{light green} to \cw{tgBlue}{dark blue} shows $r^\ast=1-2p-2q$, the fastest decay of $\beta_n\asymp n^{-r}$ that still permits detection. The thin contour lines are the \tg\ boundaries $2p+2q+r=1$ for the labeled values of $r>0$; $T_n^{\tg}$ attains each of them, and each is itself detectable when $q,r>0$. The \cw{tgYellow}{yellow} region is detectable for Gumbel-max and maximal coupling but not for \tg, for any $r\ge0$. The \cw{tgGrey}{grey} hatched region is undetectable for all three schemes.}
\label{fig:phase_fixed}
\end{minipage}\hfill
\begin{minipage}[t]{0.48\textwidth}
\vspace{0pt}
\centering
\includegraphics[width=\linewidth]{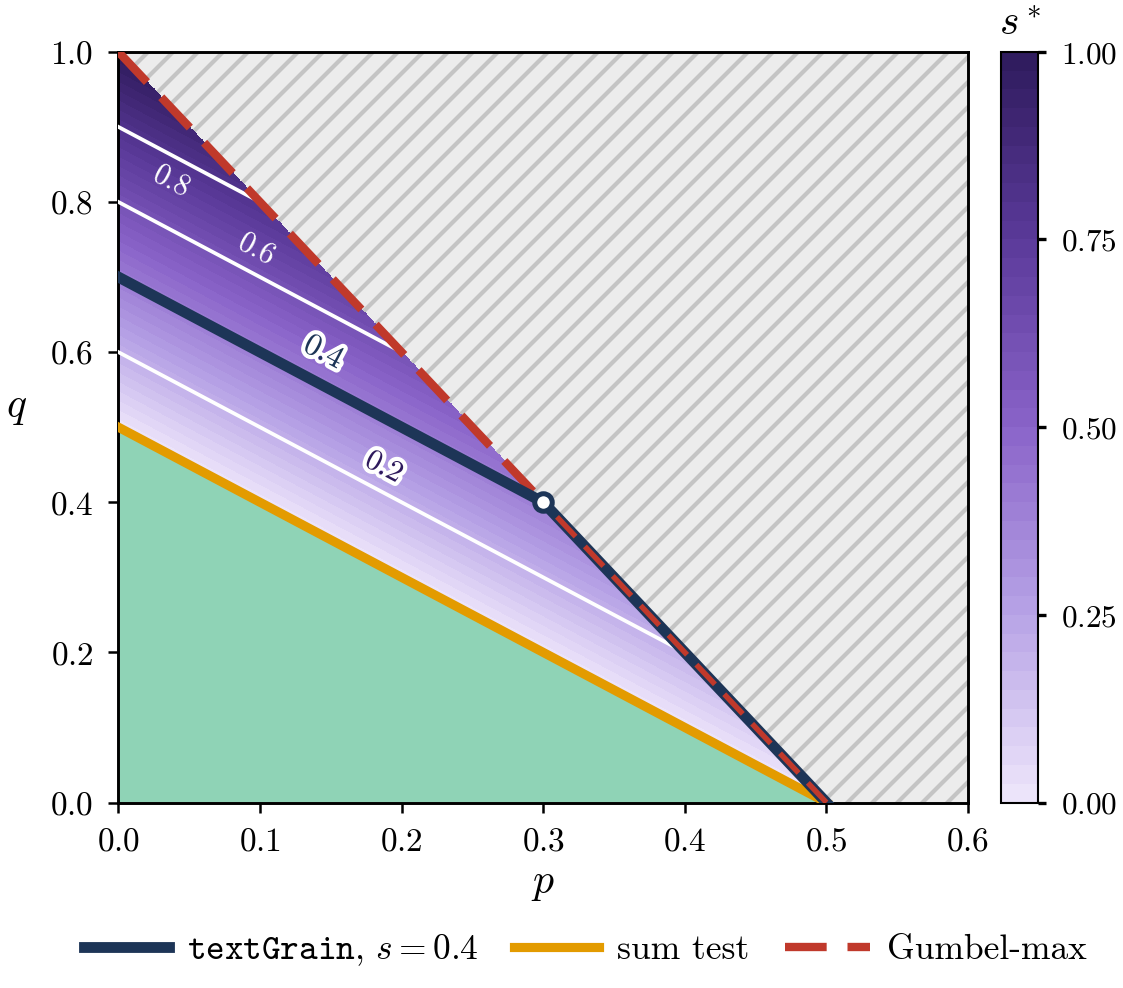}
\caption{\small Phase diagram for \tg\ with $m_n\asymp n^{s}$ columns and $\beta_t\ge\beta_0$. In the \cw{tgGreen}{green} region below the \cw{tgGold}{gold} line $p+q=1/2$, both $T_n^{\tg}$ and Algorithm~\ref{alg:test} are consistent. Between the \cw{tgGold}{gold} line and the \cw{tgRed}{red} dashed line $2p+q=1$, the shading from \cw{tgLavender}{lavender} to \cw{tgPurple}{dark purple} shows $s^\ast=2p+2q-1$. Consistent tests exist if and only if $s>s^\ast$ (Theorem~\ref{thm:activehigh}), Algorithm~\ref{alg:test} attains them (Theorem~\ref{thm:activehigh_adaptive}), and $T_n^{\tg}$ is powerless there (Proposition~\ref{prop:sumloss}). Each contour line, continued along the \cw{tgRed}{red} line below $q=s$, is the \tg\ boundary $2p+2q-\min(s,q)=1$ for its labeled value of $s$. The \cw{tgNavy}{navy} line draws one such boundary in full, for the illustrative value $s=0.4$, with the circle marking its kink at $q=s$. The \cw{tgGrey}{grey} hatched region is undetectable for every $s$. Under a weak entropy budget (Theorem~\ref{thm:activelow}), the boundary does not depend on $s$, Figure~\ref{fig:phase_fixed} applies, and $T_n^{\tg}$ remains optimal (Theorem~\ref{thm:activelow_sum}).}
\label{fig:phase_growing}
\end{minipage}
\end{figure}

\paragraph{Our Contributions.} Our contributions are summarized as follows.
\begin{itemize}
    \item  \textbf{\tg\ makes detection harder:} We establish that, with a fixed number of columns and entropy fractions $\beta_n\asymp n^{-r}$, the minimax detection boundary of \tg\ is $2p+2q+r=1$, which reduces to $p+q=1/2$ when the entropy fraction is bounded away from zero (Theorem~\ref{thm:boundary} and Corollary~\ref{cor:phase}). The detection boundary for both the Gumbel-max watermarking  \citep[Theorem 2]{li2026robust} and the maximal coupling watermarking \citep[Theorem 3.5]{xie2025debiasing} is given by $2p+q=1$. Therefore, the detectable region for the minimax optimal detector of the \tg\ watermark is strictly contained inside the detectable region of Gumbel-max or maximal coupling-based watermarks; hence, \tg\ watermarks are strictly harder to detect. A richer phase transition beyond merely Theorem \ref{inf-thm-boundary}, with $\beta_n \asymp n^{-r}$, is presented in Figure \ref{fig:phase_fixed} or Table \ref{tab:comp}, as well as in Theorem \ref{thm:boundary}.
    
    \item \textbf{Sum test is consistent throughout the detectable regime when the number of columns is fixed.} We show that, when the number of columns is fixed, the report's statistic $T_n^\tg$ attains this minimax boundary for the testing problem \eqref{eq:mixture}, and is therefore consistent whenever detection is possible. This is mostly an artifact of \tg\ making watermark detection harder: for the maximal-coupling watermark, the sum test of \citet{xie2025debiasing} also has boundary $p+q=1/2$, which there renders it suboptimal. This optimality, however, does not extend to a growing number of columns, where $T_n^\tg$ can miss detectable alternatives. A detailed exposition can be found in \S\ref{sec:sum-opt}.
    
    \item \textbf{Sum test is suboptimal when number of columns grows.} In Proposition \ref{prop:sumloss}, we show that when we allow a large entropy budget as well as growing $m_n$, $T_n^\tg$ fails to achieve the optimal detection boundary; see Figure \ref{fig:phase_growing} for the corresponding phase-diagram with $m_n \asymp n^s$, $\varepsilon_n \asymp n^{-p}$, and $\Delta_n \asymp n^{-q}$. Moreover, even where $T_n^\tg$ attains the detection boundary, it is never exponentially efficient. Following the efficiency criterion of \citet{li2026robust}, we identify the optimal exponential rate at which the type II error of any level-$\alpha$ detector can decay under \tg\ (Theorem~\ref{thm:efficiency}). We then show that, at every fixed \tg\ alternative, the type II error of $T_n^\tg$ decays at a strictly slower rate (Proposition~\ref{prop:effsum}).

    \item \textbf{An adaptive test that is optimal in both senses.} To remedy these shortcomings, we propose a weighted multiscale binomial test (Algorithm~\ref{alg:test}) that uses the same token scores as $T_n^\tg$ and requires no knowledge of the NTPs, the entropy budget, or the edit fraction. It attains every detection boundary established in this paper, including those missed by $T_n^\tg$ (Theorems~\ref{thm:boundary_adaptive} and~\ref{thm:activehigh_adaptive}), as well as the optimal exponential rate (Theorem~\ref{thm:efficiency_adaptive}).
\end{itemize}

\begin{table}[h!]
\centering
\small
\caption{\small Detection boundaries in the $(p,q)$ plane, where $\varepsilon_n\asymp n^{-p}$, $\Delta_n\asymp n^{-q}$ and, for growing $m$, $m_n\asymp n^{s}$. Both \tg\ rows assume entropy fractions $\beta_n \gtrsim 1$; no test separates the hypotheses above the boundary, and consistent tests exist below it.}
\label{tab:comp}
\begin{tabular}{@{}ll@{}}
\toprule
Watermark & Detection boundary\\
\midrule
\tg, fixed $m$ & $2p+2q=1$\\[2pt]
\tg, growing $m$ & $2p+2q-\min(s,q)=1$\\[2pt]
Gumbel-max & $2p+q=1$\\[2pt]
Maximal coupling$^\dagger$ & $2p+q=1$\\
\bottomrule
\end{tabular}

\vspace{4pt}
\parbox{0.62\linewidth}{\footnotesize $^\dagger$\citet{xie2025debiasing} define $\Delta(P_t)=1-P_{t,G_t}$ rather than $1-\max_w P_{t,w}$; see Remark~\ref{rem:finite_comparison}.}
\end{table}

We end this section with an interesting observation. In contrast to our intuition, the shrinking detection boundary in Theorem \ref{inf-thm-boundary} seems to be not merely an artifact of retaining randomness- in fact, maintaining entropy-calibration in an watermark scheme need not lead to harder watermark detection. In particular, in Example \ref{ex:matched-entropy-separation}, we construct a toy unbiased and entropy-preserving watermarking scheme $\mathrm{RG}$ such that $H_0$ and the corresponding alternative $H_1^{\mathrm{RG}}$ separates in a regime where $H_0$ and $H_1^{\tg}$ asymptotically merges into one another.  Consider the following example.
 \begin{example}
 In this example, we construct a toy watermarking scheme $\mathrm{RG}$ (\textbf{R}andomized \textbf{G}umbel-max) which itself is unbiased and entropy-calibrated, and show that there exists a regime where the hypotheses corresponding to $\mathrm{RG}$ asymptotically separates, whereas those corresponding to \tg\ do not.
\label{ex:matched-entropy-separation}
Set $B=m=2$. For $0<\delta<1/4$, put
$P_{\delta,w_0}=1-\delta$ and $P_{\delta,w_i}=\delta/ |\cW-1|$, 
$i\ge1$. Consider a fixed partition $g(w_0)=1$ and $g(w_i)=2$
for $i\ge1$. Thus $P_\delta$ has full token support and
$P_\delta^{\mathrm{blk}}=(1-\delta,\delta)$.
Fix $c\in(0,\log2)$ and set
$\beta_\delta=c\delta/H(P_\delta)$. Define
\[
 b^*(Z,J)=\mathop{\mathrm{argmax}}_{b\in[2]}
 \{\log P_{\delta,b}^{\mathrm{blk}}+Z_{bJ}\}.
\]
For each $u\in (0,1)$, let $\IP_u$ select block $b=2$ with probability $u$. Let $\mathbb{H}(u)=H(\IP_u)$ and, let $a_\delta\in(0,1)$ satisfy
\[
\begin{split}
c\delta = \mathbb{H}(\delta)
 -(1-\delta)\mathbb{H}((1-a)\delta) -\delta \mathbb{H}(a+(1-a)\delta).
\end{split}
\]
Define the randomized Gumbel-max
conditional token distribution by
\begin{align}
    Q^{\mathrm{RG}}_\delta(w\mid Z,J)
 =\left[(1-a_\delta)P_{\delta,g(w)}^{\mathrm{blk}}
 +a_\delta\mathbf1\{b^*(Z,J)=g(w)\}\right]
 \frac{P_{\delta,w}}{P_{\delta,g(w)}^{\mathrm{blk}}}. \label{eq:new-wm}
\end{align}
It is easy to verify that for all sifficiently small $\delta>0$, \eqref{eq:new-wm} defines an unbiased watermark scheme satisfying $ H(P_\delta)-\mathbb E_{Z,J}H\{Q^k_\delta(\cdot\mid Z,J)\}
 =c\delta.$ Consider as $H_1^{\mathrm{RG}}$ the hypothesis corresponding to $H_1^\tg$ in \eqref{eq:mixture}. Then, taking $\Delta_n=n^{-2/3}$ and $\varepsilon_n=1$ leads to a detectable separation between $H_0$ and $H_1^{\mathrm{RG}}$:
\[
\begin{aligned}
 \inf_{\phi_n}\{\mathbb E_0\phi_n
 +\mathbb E_{\mathrm{RG},n}(1-\phi_n)\}&\to0,
\end{aligned}
\]
In contrast, for \tg\ watermarking, it is impossible to separate $H_0$ and $H_1^{\tg}$ asymptotically under the same setting.
\[ \inf_{\phi_n}\{\mathbb E_0\phi_n
 +\mathbb E_{\tg,n}(1-\phi_n)\}\to1. \]
See Proposition \ref{prop:example} for a detailed statement and the proof.  This example shows that the stricter detectable regime is not an immediate consequence of fusing entropy-calibration into unbiased watermarking; the detection boundary arises purely out of construction of \tg. 
\end{example}

\subsection{Notations}
In this paper, we denote the set $\{1, \ldots, n\}$ by $[n]$, and the $d$-dimensional
Euclidean space by $\R^d$. We write $\IP_0$ and $\E_0$ for probability and expectation
under the null hypothesis, and $\IP_1$ and $\E_1$ for probability and expectation under
the alternative; the latter are often augmented with subscripts, as in
\eqref{eq:mixture}, to indicate their dependence on the relevant quantities. For a
distribution $P$ with density (or probability mass function) $p$, we denote its entropy
by $H(P) = -\E_{X\sim P}\{\log p(X)\}$. Specifically, for $x\in[0,1]$, let
\begin{equation}
\label{eq:binary-entropy}
 \mathbb H(x)=-(1-x)\log(1-x)-x\log x
\end{equation}
denote the entropy of the two-point probability vector $(1-x,x)$, with
the convention $0\log0=0$. For non-negative sequences $(a_n)_{n \geq 1}$
and $(b_n)_{n \geq 1}$, we write $a_n \lesssim b_n$ if $a_n \le C b_n$ for all
$n \geq 1$ and some constant $C > 0$, and $a_n \asymp b_n$ if
$C_1 b_n \le a_n \le C_2 b_n$ for all $n \geq 1$ and some constants $C_1, C_2 > 0$.
We also write $a_n = O(b_n)$ to mean $a_n \lesssim b_n$, and $a_n = o(b_n)$ if
$a_n/b_n \to 0$. For densities $f, g$ on $(0,1)$, let $\chi^2(f \wall g)$ and
$\KL(f \wall g)$ denote the chi-squared and Kullback--Leibler divergences,
respectively, defined as
\[
  \chi^2(f\wall g)=\int_0^1\frac{\{f(y)-g(y)\}^2}{g(y)}\dd y,
  \qquad
  \KL(f\wall g)=\int_0^1 f(y)\log\frac{f(y)}{g(y)}\dd y,
\]
with the convention $0/0=0$, and with both divergences set equal to
$+\infty$ if $f>0=g$ on a set of positive Lebesgue measure. The maximum of an empty set is $-\infty$.

\section{Detection boundary for \tg}
\label{sec:model}
This section devotes itself to a thorough treatment of the optimal detection boundary for any potential \tg-detector. In particular, \S\ref{sec:info-asmp} introduces the key assumptions that enables a systematic treatment of the alternative hypothesis $H_1^\tg$, following which, we address the questions of detection boundary, and the optimality of the sum-test $T_n^\tg$, in the subsequent sections.

\subsection{Preliminaries: assumptions on \tg and watermark detection}\label{sec:info-asmp}
At any position, the \tg\ watermark has five key components: NTP $P$, the partitions $g$, the pseudo-random variables $Z_{bj}$, the random draw $J\sim \Unif([m])$, and the entropy regularizer $\beta$ \citep{textgrain2026}. 
Assumption \ref{ass:experiment} rules out pathological schemes by explicitly specifying the inter-dependence of all these components. Precisely, 
if  $\cF_t$ is the filtration containing the generation and key history
through position $t$, then we presume the following. 
\begin{assumption}
\label{ass:experiment}
At position $t$, $P_t$ and $\beta_t\in[0,1]$ are
$\cF_{t-1}$-measurable. The partition \(g_t\), column $J_t$ and pseudo-random variables $\bz$ are drawn independently at every step $t$.
\end{assumption}
Assumption~\ref{ass:experiment} retains the conditional mixture structure
of \citet[Assumption 1]{li2026robust}. In particular, under the null of no watermarking, the random draws used for the partition $g$, pseudo-randomness $\bz$, and selected column $J$, are independent at a fixed context. This, in turn, makes the NTP $P_t$ and the pivot statistics $Y_t:=Z_{t, g(W_t), J_t}\sim \Xi$ independent. In contrast, under watermarking, they are obviously not independent; the parameter $\beta$ dictates the strength of \tg\ watermarking, i.e. how far the joint distribution $(P_t, Y_t)$ can venture out to from $P_t \otimes \Xi$. 

Note that in \tg's official implementation \cite{textgrain2026},  only the first occurrence of each context window is scored. A fixed $\mathsf{Key}$ and a fixed finite
window allow only finitely many distinct contexts, which doesn't facilitate asymptotic analysis. There are a few ways to circumvent this: for example, \cite{cai2025optimal, blanchet2026optimal} considers a growing vocabulary size; on the other hand, one may also consider a growing context space or a fresh $\mathsf{Key}$ to ensure randomness. As a convention, we presume the latter: each scored position uses a fresh keyed
randomness conditional on the past. Accordingly, all asymptotics are taken as
$n\to\infty$, where $n$ is a deterministic number of fresh scored
positions.

\noindent
We also need to take care of the stochasticity arising out of the partition $g$. Assumption \ref{ass:partition} specifies when token uncertainty remains visible after
partitioning.
\begin{assumption}
\label{ass:partition}
There is a constant $\vartheta>0$ such that, at every position and for
every pair of distinct tokens $w,v\in\cW$,
$\Pp\{g_t(w)\ne g_t(v)\mid\cF_{t-1}\}\ge\vartheta$ almost surely.
\end{assumption}
In particular, independent uniform assignments to $B$ blocks satisfy
Assumption~\ref{ass:partition} with $\vartheta=1-1/B$. Nevertheless, we emphasize that Assumption \ref{ass:partition} allows for more general choices of partitions.  Before moving on to the detection boundary, we round off this discussion by specifying what we mean as a watermark detector. 

\paragraph{How would a generic \tg\ detector work?}
A \tg\ detector would primarily consist of a \emph{score test} $\phi_n:[0,1]^n\to[0,1]$, which may employ the scored tokens $\{w_t\}_{t\in [n]}$, partitions $\{g_t\}$, the corresponding variables $Y_t$, and selected columns $J_t$.  In general, the pivot distributions $(P_t)$ is not available to the detector. For the analysis of an arbitrary detector (we also use the phrase "score test" interchangeably), we characterize the general class of alternatives for the test \eqref{eq:mixture}.
\begin{definition}[Token alternative class and minimax total error]
\label{def:class}
Let
$\cH_{1,n}(\Delta_n,\beta_n)$ be the nonempty class of alternative
laws satisfying Assumptions~\ref{ass:experiment} and 
\ref{ass:partition}. In particular, the distributions $P$ in $\cH_{1,n}(\Delta_n,\beta_n)$ satisfies 
\[\Delta_n\le\Delta(P_t)\le2\Delta_n,
 \ 
 \beta_n\le\beta_t\le2\beta_n.\]
Define the corresponding minimax risk on this class over the score tests as,
\begin{equation}
    \label{eq:minimax risk}
     R_n^Y=\inf_{\phi_n:[0,1]^n\to[0,1]}
 \left\{\E_0\phi_n+
   \sup_{\Pp_1\in\cH_{1,n}(\Delta_n,\beta_n)}
             \E_1(1-\phi_n)\right\}.
\end{equation}
\end{definition}

In the following section, we proceed towards identifying the optimal detection boundary for \tg\ via a minimax analysis of $R_n^Y$.

\subsection{Theoretical guarantees: detectability}
\label{sec:information}
Before embarking on an analysis of the detector, we formalize the intuition that \tg\ works by providing a signal analysis. To that end, recall $f_{1, P, \beta, g}$; let $\Pp_{1,P,\beta,g}$ denote the corresponding alternative law, so that $Y=F_G(Z_{g(W),J}) \sim \Pp_{1,P,\beta,g}$. Note that \begin{equation}
\label{eq:selected_score_density}
 f_{1,P,\beta,g}(y)
 =\sum_{b=1}^B\sum_{j=1}^m
   \E\!\left[
      \pi_{P,\beta}(b,j)\,\middle|\,
      Z_{bj}=-\log(-\log y)
   \right],\  0<y<1.
\end{equation}
Theorem \ref{cor:tokens} characterizes the signal strength for any natural score function $h$, applied on the pivot statistics $Y_t$ from \eqref{eq:score_convention}.
\begin{theorem}[Token-level signal strength]
\label{cor:tokens}
Grant Assumption \ref{ass:experiment}. Let $m\ge2$, and let $g$ satisfy Assumption~\ref{ass:partition}.
Set $f_{1,P,\beta}=\E_gf_{1,P,\beta,g}$, and let
$\E_{1,P,\beta}$ denote expectations under 
$Y\sim f_{1,P,\beta}$. Then $\chi^2(f_{1,P, \beta} \wall \Unif[0,1])
 \asymp_{m,\vartheta}
 \Delta(P)\min\{\Delta(P),\beta H(P)\}.$ 
In particular, for every fixed nondecreasing $h\in L^2(0,1)$ with
$\Var_0\{h(Y)\}>0$,
\[
 \E_{1,P,\beta}h(Y)-\E_0h(Y)
 \asymp_{m,\vartheta,h}
 \sqrt{\Delta(P)\min\{\Delta(P),\beta H(P)\}}.
\]
Consequently, this holds almost surely conditional on
$\cF_{t-1}$.
\end{theorem}
Note that \cite{textgrain2026} suggests $h(y)= - \log(1-y)$; the same conclusion holds for this choice as well. 
\begin{remark}
    Theorem \ref{cor:tokens} hints at an interesting interplay between hardness of a detection problem, and the entropy calibrated by \tg. Moreover, observing that
\[
H(P)\ge -(1-\Delta(P))\log(1-\Delta(P))-\Delta(P)\log\Delta(P)
\ge \Delta(P)\log(1/\Delta(P)),
\]
it follows that if $\beta \asymp 1$,
$H(P)/\Delta(P)\ge\beta\log(1/\Delta(P))\to\infty$ as
$\Delta(P)\downarrow0$. Thus, from Theorem \ref{cor:tokens},  \[\E_{1,P,\beta}h(Y)-\E_0h(Y)
 \asymp_{m,\vartheta,h} \Delta(P) \]
 for all sufficiently small $\Delta(P)$. For comparison, consider the Gumbel-max watermark and write
$\E^{\mathrm{GM}}_{1,P}$ for expectation under its watermarked
pivot distribution. For the same score $h(y)=-\log(1-y)$,
\citet[Proposition A.1 in the supplementary material]{li2026robust}
show that $\E^{\mathrm{GM}}_{1,P}h(Y)-\E_0h(Y) \asymp H(P)$.
Combining these bounds with Theorem \ref{cor:tokens} gives
\[
\frac{\E_{1,P,\beta}h(Y)-\E_0h(Y)}
{\E^{\mathrm{GM}}_{1,P}h(Y)-\E_0h(Y)}
\asymp_{m,\vartheta}\frac{\Delta(P)}{H(P)}
\longrightarrow0, \ \text{as } \Delta(P) \to 0.
\]
For a fixed vocabulary,
$H(P)\asymp\Delta(P)\log(1/\Delta(P))$.
Hence, relative to Gumbel-max, \tg\ loses a logarithmic
factor in the signal strength of the corresponding detectors.
\end{remark}

 Under $H_1^{\tg}$ in \eqref{eq:mixture}, the score at position $t$ has conditional density $1-\varepsilon_n+\varepsilon_n f_{1,t}$ given $\cF_{t-1}$, whose $\chi^2$-divergence from $\Unif(0,1)$ equals $\varepsilon_n^2\,\chi^2(f_{1,t}\wall\Unif[0,1])$. By Theorem~\ref{cor:tokens}, each position therefore contributes information of order $\varepsilon_n^2\Delta(P_t)\min\{\Delta(P_t),\beta_tH(P_t)\}$, and summing over the $n$ positions suggests that in light of $H(P_t)\asymp \mathbb{H}(\Delta_n)$, detection is governed by
\[
 I_n:=n\varepsilon_n^2\Delta_n\min\{\Delta_n,\beta_n\mathbb H(\Delta_n)\}.
\]

Theorem \ref{thm:boundary} formalizes the above intuition.

\begin{theorem}[Detection boundary]
\label{thm:boundary}
 Recall $R_n^Y$ from Definition~\ref{def:class}. Let $m\ge2$ be
fixed, and suppose Assumptions~\ref{ass:experiment} and
\ref{ass:partition} hold.
\begin{itemize}
\item  If
$\sup_n n\varepsilon_n^2\Delta_n
\min\{\Delta_n,\beta_n\mathbb H(\Delta_n)\}<\infty$, then
$\liminf_{n\to\infty}R_n^Y>0$. 
    \item If
$n\varepsilon_n^2\Delta_n
\min\{\Delta_n,\beta_n\mathbb H(\Delta_n)\}\to0$, then
$\sup_{\Pp_1\in\cH_{1,n}(\Delta_n,\beta_n)}\TV(\Pp_0,\Pp_1^Y)\to0$
and $R_n^Y\to1$.

\item Finally $R_n^Y\to0$ if and only if
$n\varepsilon_n^2\Delta_n
\min\{\Delta_n,\beta_n\mathbb H(\Delta_n)\}\to\infty$.
\end{itemize}

\end{theorem}

The polynomial specialization gives an explicit boundary in the
sparsity, token uncertainty, and entropy-fraction exponents.

\begin{corollary}[Phase diagram]
\label{cor:phase}
Grant the conditions and notations of Theorem \ref{thm:boundary}. Suppose $\varepsilon_n\asymp n^{-p}$, $\Delta_n\asymp n^{-q}$, and
$\beta_n\asymp n^{-r}$ for fixed $p,q,r\ge0$.
\begin{itemize}
\item  If $2p+2q+r<1$, or if
$2p+2q+r=1$ with $q,r>0$, then $R_n^Y\to0$
\item If $2p+2q+r>1$, then
$\sup_{\Pp_1\in\cH_{1,n}(\Delta_n,\beta_n)}\TV(\Pp_0,\Pp_1^Y)\to0$
and $R_n^Y\to1$.
\item If $2p+2q+r=1$ and $\min(q,r)=0$, then
$\liminf_{n\to\infty}R_n^Y>0$. In particular, if $\beta_n\asymp 1$, then the detection boundary is given by
$p+q=1/2$, and $R_n^Y\not\to0$ on it.
\end{itemize}
\end{corollary}

\begin{remark}[Comparison of the finite-column boundaries]
\label{rem:finite_comparison}
If $\beta_n \asymp 1$, Corollary~\ref{cor:phase}
gives the \tg\ boundary $p+q=1/2$.
In contrast, the Gumbel-max boundary in \citet[Theorems 2--3]{li2026robust}
and the weak-signal maximal-coupling boundary in
\citet[Theorems 3.5 and 3.7]{xie2025debiasing} are $2p+q=1$. For comparison, we must note that for \cite{xie2025debiasing}, $\Delta (P_t)$ is defined as $1-P_{t,G_t}$ instead of 
$1-\max_wP_{t,w}$ for Gumbel-max or \tg. Finally, the dependence on a vanishing $\beta_n$ in Corollary~\ref{cor:phase}
also reflects \tg\'s fractional conditional-entropy constraint,
which is absent from those two weak-signal parametrizations.
\end{remark}
As mentioned before, the optimal detection boundary for \tg\ thus coincides
numerically with the sum-test boundary in \cite{li2026robust} or \cite{xie2025debiasing}. We will formally establish the analogous result for $T_n^\tg$ in \S\ref{sec:sums}. Moving forward, we investigate the detection boundaries when the number of columns $m$, is allowed to diverge.

\subsection{Detection boundary with growing columns}
\label{sec:growing}

This section devotes itself to optimal boundary detection when the number of columns, $m$, is allowed to diverge with $n$. Analogous to Definition \ref{def:class}, we consider the following class of alternatives. 

\begin{definition}\label{defn:high}
    Let
$\cH_{1,n}^{\mathrm{high}}$ be the nonempty class of alternatives, such that at every position $t$
$$\Delta_n\le\Delta(P_t)\le\min(2\Delta_n,1/4), \ \text{ and }
\beta_t \ge \beta_0$$. The corresponding minimax risk is given by 
\[
 R_n^{\mathrm{high}}:= \inf_\phi\Bigl\{\E_0\phi
       +\sup_{\Pp_1\in\cH_{1,n}^{\mathrm{high}}}\E_1(1-\phi)\Bigr\},
\]
\end{definition}

When the absolute entropy budget is large compared with the rare-block
mass, the next theorem preserves the nonbinding detection exponent.
It does not assert a constant-factor equivalence of the information in
the two budget regimes.

\begin{theorem}[strong entropy budget]
\label{thm:activehigh}
Suppose Assumptions~\ref{ass:experiment} and \ref{ass:partition} hold, and assume that
$\Delta_n\asymp n^{-q}$ and $m_n\asymp n^s$ with $q,s>0$. Let $f_{1, P_t, \beta_t}:=\E[f_{1,P_t, \beta_t, g_t}\mid \cF_{t-1}]$.
\begin{enumerate}
\item[(a)] Uniformly over $\Pp_1\in\cH_{1,n}^{\mathrm{high}}$ and $1\le t\le n$, it follows that $\chi^2(f_{1, P_t, \beta_t} \wall \Unif[0,1])
 =n^{-2q+\min(s,q)+o(1)}$.

\item[(b)]  Suppose
$\varepsilon_n\asymp n^{-p}$ with $p\ge0$, and consider the minimax risk $R_n^{\mathrm{high}}$ from Definition \ref{defn:high}.
\begin{itemize}
    \item If
$2p+2q-\min(s,q)<1$, $R_n^{\mathrm{high}}\to 0.$ 
\item If $2p+2q-\min(s,q)=1$, then $\liminf_{n\to \infty} R_n^{\mathrm{high}}>0$.
\item Finally, if $2p+2q-\min(s,q)>1$, then every score test $\psi_n:[0,1]^n\to [0,1]$ satisfies \[
 \sup_{\Pp_1\in\cH_{1,n}^{\mathrm{high}}}
 \bigl|\E_0\psi_n+\E_1(1-\psi_n)-1\bigr|\to0.
\]
\end{itemize}
\end{enumerate}
\end{theorem}
Theorem~\ref{thm:activehigh} highlights a computation-accuracy trade-off. With a strong entropy budget, the per-position information grows like $\Delta\min\{m\Delta,1\}$, so additional columns improve detectability only until $m\asymp1/\Delta$.  On the other hand, generating a \tg\ token requires sampling a $B\times m$ Gumbel table and solving \eqref{eq:transport} over it, at a cost of at least $O(Bm)$.  
\begin{remark}[The role of the entropy budget]
\label{rem:budget_role}
The entropy fraction $\beta$ interpolates between unwatermarked and fully key-dependent generation. When $\beta=0$, the constraint in \eqref{eq:transport} forces $\pi_{P,0}=P^{\blk}\otimes m^{-1}\bo$, so that the token is drawn from $P$ independently of the key, i.e. under the null. As the budget $\beta$ increases, the key exerts more influence on the token and the detectable region expands. This gain, however, saturates. Once the absolute budget dominates the rare-block mass, Theorem~\ref{thm:activehigh} shows that the boundary $2p+2q-\min(s,q)=1$ no longer depends on the budget and coincides with that of the maximal fraction $\beta=1$. Beyond this point, detectability is limited by the token uncertainty and the number of columns rather than by $\beta$. In terms of our first question, the entropy budget therefore trades preserved sampling randomness for detectability only up to the scale of the rare-block mass.
\end{remark}

The next theorem treats entropy budgets that are small relative to
block uncertainty. Fix $B\ge2$, and let $\Delta_n,\kappa_n>0$.
Let $\cH_{1,n}^{\mathrm{low}}$ be the nonempty class of alternative
laws satisfying Assumptions~\ref{ass:experiment} and
\ref{ass:partition}, such that, at every position $t$, almost surely,
\[
 \Delta_n\le\Delta(P_t^{\blk})\le\min(2\Delta_n,1/4),
  \kappa_n\le\beta_tH(P_t)\le2\kappa_n.
\]
The corresponding minimax risk is defined by
\[R_n^{\mathrm{low}}:=\inf_\phi\Bigl\{\E_0\phi
       +\sup_{\Pp_1\in\cH_{1,n}^{\mathrm{low}}}\E_1(1-\phi)\Bigr\}.\]

\begin{theorem}[Growing columns with a weak entropy budget]
\label{thm:activelow}
Consider the class $\cH_{1,n}^{\mathrm{low}}$ defined above, and suppose $\kappa_n/\Delta_n\to0$ and
$e^{-m_n/64}=o(\kappa_n)$. 
\begin{enumerate}
\item[(a)] It follows that $\chi^2(f_{1,P_t,\beta_t,g_t} \wall \Unif[0,1])\asymp_B\Delta_n \kappa_n.$ 
 \item[(b)] For every fixed nondecreasing $h\in L^2(0,1)$ with
$\Var_0\{h(Y)\}>0$,
\[
 \int_0^1h(y)\{f_{1,P_t,\beta_t,g_t}(y)-1\}\dd y
 \asymp_{B,h}\sqrt{\Delta_n \kappa_n}.
\]
\item[(c)] $R_n^{\mathrm{low}}\to 0$ if and only if $n\varepsilon_n^2\Delta_n\kappa_n\to\infty.$ Moreover, if $n\varepsilon_n^2\Delta_n\kappa_n\to0$, then every measurable score
test $0\le\psi_n\le1$ satisfies
\[
 \sup_{\Pp_1\in\cH_{1,n}^{\mathrm{low}}}
 \bigl|\E_0\psi_n+\E_1(1-\psi_n)-1\bigr|\to0.
\]
\end{enumerate}
\end{theorem}

\begin{remark}[What the entropy-budget results determine]
\label{rem:budget_comparison}
Theorem~\ref{thm:activehigh} treats fixed positive entropy
fractions and gives the strict growing-column boundary
$2p+2q-\min(s,q)=1$.
In contrast, under the block-uncertainty conditions of
Theorem~\ref{thm:activelow}, a weak entropy budget gives
information of order $\Delta_n\kappa_n$ and a mean shift of
order $\sqrt{\Delta_n\kappa_n}$ for every fixed nonconstant
nondecreasing square-integrable score.
Interestingly, this does not depend on the column
growth rate, subject to $e^{-m_n/64}=o(\kappa_n)$.
\end{remark}

\section{Optimality of test based on sum statistic \texorpdfstring{$T_n^{\texttt{textGrain}}$}{T\_n\^{}textGrain}}
\label{sec:sums}
Having characterized when detection of \tg\ is possible at all, we now turn to the second question raised in Section~\ref{se:intro}: does the sum statistic $T_n^{\tg}$ proposed by \citet{textgrain2026} detect the watermark whenever the information-theoretic limits of Section~\ref{sec:model} permit? We answer this question in two stages, first for a fixed number of columns $m$, where the answer is affirmative, and then for a growing number of columns, where the $T_n^\tg$ can miss alternatives that nevertheless remain detectable.

\subsection{Fixed-column sum statistics} \label{sec:sum-opt}

We begin with the fixed-column regime of Section~\ref{sec:information}, which corresponds to the default configuration of \tg. In this regime, Theorem~\ref{cor:tokens} already suggests that $T_n^\tg$ is consistent. To elaborate, every fixed nondecreasing score shifts its mean by an amount of order $\sqrt{\Delta(P)\min\{\Delta(P),\beta H(P)\}}$, which we refer to as \textit{information}, since this is precisely the square root of the per-token information driving the lower bound in Theorem~\ref{thm:boundary}. Summing $n$ such scores, of which a fraction $\varepsilon_n$ carry the signal, should therefore separate the two hypotheses as soon as $n\varepsilon_n^2\Delta_n\min\{\Delta_n,\beta_n\mathbb H(\Delta_n)\}$ diverges. The following theorem makes this heuristic rigorous in two forms: first for a generic monotone score with a Chebyshev-type threshold, and then for the report's statistic $T_n^{\tg}$ calibrated by its exact $\operatorname{Gamma}(n,1)$ null distribution.

\begin{theorem}
\label{thm:sums}
\begin{itemize}
    \item[(a)] Let $h\in L^2(0,1)$ be fixed
and nondecreasing, with
$0<\sigma_h^2=\Var_0\{h(Y)\}<\infty$, and $\cH_{1,n}(\Delta_n,\beta_n)$ be the class in Definition~\ref{def:class}. Following \eqref{eq:binary-entropy}, if $a_n=o\!\left(
  \sqrt{n\varepsilon_n^2\Delta_n
       \min\{\Delta_n,\beta_n\mathbb H(\Delta_n)\}}
 \right),$ and $a_n \to \infty$, then under  Assumptions~\ref{ass:experiment}, and,
\ref{ass:partition}, the test
\[
 \phi_{n,h}
 =\ind\!\left\{
    \sum_{t=1}^nh(Y_t)>
       n\int_0^1h(y)\dd y+\sigma_h\sqrt n\,a_n
 \right\}
\]
has total error satisfying
\[
 \begin{split}
 \Pp_0(\phi_{n,h}=1)
 +\sup_{\Pp_1\in\cH_{1,n}(\Delta_n,\beta_n)}
       \Pp_1(\phi_{n,h}=0)  =
 O_{m,\vartheta,V,h}\!\left(
   a_n^{-2}
   +\frac1{n\varepsilon_n^2\Delta_n
             \min\{\Delta_n,\beta_n\mathbb H(\Delta_n)\}}
 \right)
 \to0.
 \end{split}
\]
\item[(b)] For $T_n^{\tg}$ in \eqref{eq:score_convention}, let $c_{n,\alpha}$ denote the
$(1-\alpha)$-quantile of the $\operatorname{Gamma}(n,1)$ law.
Fix $\alpha_0\in(0,1)$ and let
$\alpha_n=\alpha_0/\log(n+e)$.
If
$n\varepsilon_n^2\Delta_n
\min\{\Delta_n,\beta_n\mathbb H(\Delta_n)\}/\log\log(n+e)\to\infty$, then
\[
 \Pp_0(T_n^{\tg}>c_{n,\alpha_n})
 +\sup_{\Pp_1\in\cH_{1,n}(\Delta_n,\beta_n)}
       \Pp_1(T_n^{\tg}\le c_{n,\alpha_n})
       \to0.
\]
\end{itemize}
\end{theorem}
\begin{remark}[Fisher's statistic and the meaning of sum-test optimality]
\label{rem:fisher}
Since $p_t=1-Y_t$, the \cite{textgrain2026}'s detector $2T_n^{\tg}=-2\sum_t\log p_t$ is Fisher's combination statistic. Fisher's method is known to fail in very sparse Gaussian mixtures \citep[Theorem 1.5]{donoho2004}, where the signal sits in an extreme tail that a sum dilutes. Interestingly, Theorem~\ref{thm:sums} shows that this failure does not occur for \tg\ with a fixed number of columns, since, Theorem~\ref{cor:tokens} instructs that the mean shift of every monotone score is already of the order of the square root of the information. When the number of columns grows, the signal can concentrate in a small upper tail, and Proposition~\ref{prop:sumloss} shows that the $T_n^\tg$ then loses detectable alternatives. 
\end{remark}

We pause to ponder over the implications of Theorem \ref{thm:sums}. 
\begin{itemize}
    \item (a) shows that a sum test built from any fixed monotone score is consistent under exactly the condition $n\varepsilon_n^2\Delta_n\min\{\Delta_n,\beta_n\mathbb H(\Delta_n)\}\to\infty$, which Theorem~\ref{thm:boundary} identifies as necessary for every score test.
    \item (b) shows that the report's $T_n^{\tg}$, with its exact Gamma calibration and a slowly vanishing level, inherits this guarantee at the cost of only an additional $\log\log n$ factor, and therefore attains every consistency region of Corollary~\ref{cor:phase}. 
\end{itemize}
To summarize, Theorem~\ref{thm:sums} thus answers our second question in the affirmative whenever the number of columns is fixed. This stands in sharp contrast with Gumbel-max, for which sum tests are strictly suboptimal \citep[Theorem 4]{li2026robust}. The contrast is a direct consequence of our first finding: by making detection harder, \tg\ shrinks its minimax boundary from $2p+q=1$ down to the sum-test boundary $p+q=1/2$, leaving no room for a more sophisticated detector to improve on the statistic $T_n^\tg$ from \cite{textgrain2026}. 

\subsection{A detection-boundary loss for $T_n^\tg$}

We next ask whether the optimality of the $T_n^\tg$ survives once the number of columns grows. A weak entropy budget, with $\kappa_n/\Delta_n\to0$, is the natural first case. Recall that the weak entropy result in \S\ref{sec:model}, Theorem~\ref{thm:activelow}(b), shows that the mean shift of every monotone score retains the order of the square root of the information. The budget therefore keeps the signal from concentrating in a small tail, and $T_n^\tg$ attains the weak-budget boundary under the column-growth condition of Theorem~\ref{thm:activelow}. We formalize this in Theorem \ref{thm:activelow_sum}.
\begin{theorem}
\label{thm:activelow_sum}
Consider the setup of Theorem~\ref{thm:activelow}.
Fix $\alpha_0\in(0,1)$ and put $\alpha_n=\alpha_0/\log(n+e)$.
Let $c_{n,\alpha_n}$ be the $1-\alpha_n$ quantile of
$\operatorname{Gamma}(n,1)$. If
$n\varepsilon_n^2\Delta_n\kappa_n/\log\log(n+e)\to\infty$, then
\[
 \Pp_0(T_n^{\tg}>c_{n,\alpha_n})
 +\sup_{\Pp_1\in\cH_{1,n}^{\mathrm{low}}}
       \Pp_1(T_n^{\tg}\le c_{n,\alpha_n})\to0.
\]
\end{theorem}
In the above result, the condition $n\varepsilon_n^2\Delta_n\kappa_n/\log\log(n+e)\to\infty$ matches the minimax characterization of Theorem~\ref{thm:activelow}(c) up to a $\log\log n$ factor, and it does not depend on how fast $m_n$ grows. Thus, with respect to our second question, a growing number of columns does not by itself make the $T_n^\tg$ suboptimal. As we show next, the loss arises only when the entropy budget is large enough to let the signal concentrate in a small upper tail.

 When the entropy budget is large compared with the rare-block mass, Theorem~\ref{thm:activehigh} shows that a growing number of columns enlarges the detectable region to $2p+2q-\min(s,q)<1$. The next proposition shows that the $T_n^\tg$ cannot exploit this gain: its sensitivity is governed by $\varepsilon_n\Delta_n\log(e/\Delta_n)$ regardless of the number of columns, so every threshold fails once $p+q>1/2$.

\begin{proposition}[Suboptimality of thresholds for the $T_n^\tg$]
\label{prop:sumloss}
Suppose Assumption~\ref{ass:experiment} holds,
and $\Delta_n\in(0,1/4]$. Consider \eqref{eq:mixture}, with
$Y_1,\ldots,Y_n$ independent under each alternative law. Suppose
that, at every position $t$, $\Delta(P_t^{\blk})\le2\Delta_n$,
where $P_t^\blk= \sum_{w \in g(w_t)} P_w$. 
\begin{itemize}
    \item If
$\sqrt n\,\varepsilon_n\Delta_n\log(e/\Delta_n)\to0$, then every
deterministic sequence $c_n\in\R$ satisfies
\[
 \sup_{\Pp_1}\bigl|\Pp_0(T_n^{\tg}>c_n)+\Pp_1(T_n^{\tg}\le c_n)-1\bigr|
 \to0.
\]
\item Suppose that $\varepsilon_n\asymp n^{-p}$,
$\Delta_n\asymp n^{-q}$, and $m_n\asymp n^s$, with
$p\ge0$, $q,s>0$,
$p+q>1/2$, and $2p+2q-\min(s,q)<1$.
Then the preceding conclusion holds uniformly over the
nonempty subclass of $\cH_{1,n}^{\mathrm{high}}$ with
deterministic parameter sequences and prescribed partition
laws, for which the pivots are independent.
In contrast, Theorem~\ref{thm:activehigh} implies that the
minimax total error over this subclass tends to zero.
\end{itemize}

\end{proposition}

Proposition \ref{prop:sumloss} dictates that $T_n^\tg$ is asymptotically powerless for every choice of threshold whenever $p+q>1/2$, although Theorem~\ref{thm:activehigh} guarantees consistent tests throughout $2p+2q-\min(s,q)<1$. With growing columns and a large entropy budget, the answer to our second question is therefore negative: the region $\{p+q>1/2,\ 2p+2q-\min(s,q)<1\}$, which is nonempty whenever $q,s>0$, is detectable but missed by $T_n^{\tg}$. Together with Theorems~\ref{thm:sums} and~\ref{thm:activelow_sum}, this completely delineates when the report's statistic is optimal and when it's not. Additionally, it motivates the requirement of an adaptive test which attains the full detectable region. We construct such a test in Section~\ref{subsec:test}.

\section{Optimal exponential efficiency and efficient adaptive tests}
\label{sec:efficiency}
The detection boundary describes when the errors of a test vanish, but not how fast. We now address our third question: following the efficiency criterion of \citet{li2026robust}, we identify the optimal exponential rate at which the type II error of a level-$\alpha$ test can decay under \tg, ask whether the $T_n^\tg$ attains it, and construct an adaptive test that does.

\subsection{A class-dependent result for nonstationary alternatives}
Since the generation parameters $(P_t,\beta_t)$ may vary arbitrarily across positions, the relevant rate is determined by the worst case over a class of edited score densities. We begin by specifying this class.

\begin{definition}[Convex family for the efficiency comparison]
\label{def:efficiency-class}
Fix a finite vocabulary $\cW$, positive integers $B,m$, a prescribed
partition law, and $\varepsilon\in(0,1]$. For a nonempty collection
$\Theta$ of pairs $(P,\beta)$ with $\beta\in[0,1]$, define
\[
 \mathcal Q_\Theta
 =\overline{\operatorname{conv}}^{L^1(0,1)}
   \{1-\varepsilon+\varepsilon f_{1,P,\beta}:
                                      (P,\beta)\in\Theta\}.
\]
The densities $f_{1,P,\beta}$ are averaged over the prescribed partition
law. The bar denotes the closure in $L^1(0,1)$ of all finite convex
combinations of the displayed densities.
\end{definition}
The closed convex hull is needed because the parameters may change from position to position in ways the detector cannot observe. Averaging over such variation produces mixtures of the individual score densities, and the least-favourable alternative may be one of these mixtures rather than a density generated by a single pair $(P,\beta)$. The following assumption formalizes this setting: it plays the role of Definition~\ref{def:class} for the efficiency analysis, allowing $(P_t,\beta_t)$ to be chosen adaptively from $\Theta$ at every position.

\begin{assumption}[Parameter class for exponential efficiency]
\label{ass:dense}
Assumptions~\ref{ass:experiment}  hold with a fixed
$\varepsilon\in(0,1]$, a fixed finite vocabulary $\cW$, fixed $B,m$,
and the same prescribed partition law at every fresh position.
For every $t$, the $\cF_{t-1}$-measurable pair $(P_t,\beta_t)$ belongs
to the nonempty class $\Theta$ in Definition~\ref{def:efficiency-class}.
Write $\cH_{1,n}(\Theta)$ for the class of all resulting alternative
laws. This class includes every deterministic sequence of $n$ pairs in
$\Theta$.
\end{assumption}

Unlike the sparse regime of Sections~\ref{sec:model} and~\ref{sec:sums}, the edit fraction $\varepsilon$ is now held fixed, so that the errors of a good test decay exponentially in $n$. Following \citet[Definition 1 and Theorem 5]{li2026robust}, we measure efficiency by the exponential rate of the worst-case type II error at a prescribed significance level. The next theorem identifies the optimal rate over $\cH_{1,n}(\Theta)$ and shows that it is attainable.


\begin{theorem}[Optimal class-dependent exponential efficiency]
\label{thm:efficiency}
Under Assumption~\ref{ass:dense}, every sequence of score tests
$\psi_n$ satisfying $\limsup_n\E_0\psi_n<1$ obeys
\[
 \limsup_{n\to\infty}-\frac1n\log
 \sup_{\Pp_1\in\cH_{1,n}(\Theta)}\E_1(1-\psi_n)
 \overset{(a)}{\le}
 \inf_{q\in\mathcal Q_\Theta}
 \left\{-\int_0^1\log q(y)\dd y\right\}
 <\infty.
\]
Under Assumption~\ref{ass:partition}, the infimum is strictly positive if $m\ge2$, and if there exist
$\Delta_0,\beta_0>0$ such that
$\Delta(P)\ge\Delta_0$ and $\beta\ge\beta_0$
for every $(P,\beta)\in\Theta$. Finally, under these additional conditions, if $\alpha_n\to0$ with
$\log(1/\alpha_n)=o(n)$, then there exists level-$\alpha_n$ valid tests $\{\psi_n^\star\}$ such that the equality in (a) holds. Furthermore, 
these tests $\psi_n^*$ above satisfy
\[
 \E_0\psi_n^*
 +\sup_{\Pp_1\in\cH_{1,n}(\Theta)}\E_1(1-\psi_n^*)\to0.
\]
\end{theorem}

\begin{remark}[Proof strategy, and comparison with the literature]
\label{rem:efficiency_literature}
We include a brief window into our proof strategy; detailed proofs would be included in a forthcoming extended draft. \citet[Theorem 3.2]{li2025framework} identify an optimal Gumbel-max
score and least-favourable NTP vector through a scheme-specific
convexity argument. On the other hand, \citet[Theorem 5]{li2026robust} prove that
Tr-GoF attains the corresponding mixture exponent without knowing
the edit fraction or regularity. That theorem uses a divergence index
in $(0,1)$ and no stability truncation.
In contrast, for \tg, Theorem~\ref{thm:efficiency} takes the least-favourable
comparison over the closed convex family $\mathcal Q_\Theta$ of
exact edited score laws in Definition~\ref{def:efficiency-class};
it does not assume that one NTP distribution generates the minimizing
density.
\end{remark}

In Theorem \ref{thm:efficiency}, the optimal exponent $\inf_{q\in\mathcal Q_\Theta}\{-\int_0^1\log q(y)\dd y\}=\inf_{q\in\mathcal Q_\Theta}\KL(\Unif(0,1)\wall q)$ is the Chernoff--Stein exponent against the least-favourable edited score law, and the converse shows that no score test, whatever its construction, can exceed it. The positivity statement ensures that, under nondegenerate token uncertainty and entropy fractions, \tg\ remains exponentially detectable at every fixed edit fraction. With respect to our third question, Theorem~\ref{thm:efficiency} supplies the benchmark against which any \tg\ detector should be judged. We next show that the $T_n^\tg$ falls strictly short of it.

\begin{proposition}[Strict efficiency loss of the exponential-score sum]
\label{prop:effsum}
Let $q$ be a probability density on $(0,1)$ with a nondecreasing
version, such that $\eta\le q\le M$ almost everywhere for constants
$0<\eta\le M<\infty$ and $q\ne1$ on a set of positive Lebesgue measure.
Let $\Pp_0$ be as in Theorem~\ref{thm:boundary}, and let $\Pp_q$ be the
law under which $Y_1,\ldots,Y_n$ are iid with density $q$. Fix
$\alpha\in(0,1)$, let $c_{n,\alpha}$ be the $(1-\alpha)$-quantile of
$\operatorname{Gamma}(n,1)$, and let $T_n^{\tg}$ be as in
\eqref{eq:score_convention}, so that
$\Pp_0(T_n^{\tg}>c_{n,\alpha})=\alpha$.
\begin{enumerate}
\item[(a)] \textup{(Exponent of the sum test.)} The type II error of
$\ind\{T_n^{\tg}>c_{n,\alpha}\}$ satisfies
\[
 \lim_{n\to\infty}-\frac1n\log\Pp_q(T_n^{\tg}\le c_{n,\alpha})
 =\sup_{\lambda\ge0}\Bigl\{-\lambda-
        \log\int_0^1(1-y)^\lambda q(y)\dd y\Bigr\},
\]
and
\[
 0<\sup_{\lambda\ge0}\Bigl\{-\lambda-
        \log\int_0^1(1-y)^\lambda q(y)\dd y\Bigr\}
 <-\int_0^1\log q(y)\dd y<\infty.
\]

\item[(b)] \textup{(Optimal exponent.)} There exist score tests
$\psi_n^*$ with $\E_0\psi_n^*\le\alpha$ for every $n$ such that
\[
 \lim_{n\to\infty}-\frac1n\log\E_q(1-\psi_n^*)
 =-\int_0^1\log q(y)\dd y.
\]
Conversely, every sequence of possibly randomized score tests $\psi_n$
with $\limsup_n\E_0\psi_n<1$ satisfies
\[
 \limsup_{n\to\infty}-\frac1n\log\E_q(1-\psi_n)
 \le-\int_0^1\log q(y)\dd y.
\]
\end{enumerate}
\end{proposition}

In the above result, part (a) gives the exact exponent of the report's level-$\alpha$ test, and part (b), which specializes Theorem~\ref{thm:efficiency} to a single alternative, shows that this exponent is strictly smaller than the optimal one at every non-uniform alternative. Since each $f_{1,P,\beta}$ is nondecreasing and bounded by $\min(B,m)$, the proposition applies to every \tg\ mixture $q=1-\varepsilon+\varepsilon f_{1,P,\beta}$ with $0<\varepsilon<1$ and $f_{1,P,\beta}\ne1$, with $\eta=1-\varepsilon$ and $M=1-\varepsilon+\varepsilon\min(B,m)$. Thus, with respect to our third question, the report's statistic, although optimal at the detection boundary when $m$ is fixed (Theorem~\ref{thm:sums}), is never exponentially efficient: at every fixed alternative, its type II error decays strictly more slowly than is attainable. This gap is invisible at the level of detection boundaries, and it motivates the adaptive test of Section~\ref{subsec:test}, which attains the optimal exponent.

\subsection{A weighted multiscale binomial test}
\label{subsec:test}

Propositions~\ref{prop:sumloss} and~\ref{prop:effsum} expose two distinct shortcomings of the $T_n^\tg$: it misses detectable alternatives whose signal concentrates in a small upper tail, and it loses the optimal exponential rate even when the signal is spread out. Both stem from committing to a single fixed score. We therefore propose a test that instead compares the empirical tail probabilities of the $p$-values with their null values over a grid of levels: the small dyadic levels capture concentrated signals, while a progressively finer mesh allows the test to approximate the optimal exponent without knowing the alternative.

For $n\ge1$, let $L_n=\max\{2,\lceil\log(n+1)\rceil\}$ and define the grid
\begin{equation}
\label{eq:grid}
\mathcal U_n=
\{2^{-j}:1\le j\le\lceil\log_2(n\vee2)\rceil\}
\;\cup\;\{k/L_n:1\le k<L_n\}.
\end{equation}
The dyadic levels reach down to tail probabilities of order $n^{-1}$, while the ordinary mesh becomes dense in $(0,1)$ as $n$ grows. We attach to these levels the weights
\begin{equation}
\label{eq:weights}
w_n(u)=\frac{1}{2|\mathcal U_n|}
       +\frac12\ind\{u=1/2\},\  u\in\mathcal U_n,
\end{equation}
which sum to one and reserve half of the total mass for the median, where spread-out signals are most visible. Let $\operatorname{kl}(a,u)=a\log(a/u)+(1-a)\log\{(1-a)/(1-u)\}$, for $a\in[0,1]$ and $u\in(0,1)$, denote the Bernoulli Kullback--Leibler divergence. With $p_t=1-Y_t$ as in \eqref{eq:score_convention}, let $\widehat F_n^p(u)=n^{-1}\sum_{t=1}^n\ind\{p_t\le u\}$ and define
\begin{equation}
\label{eq:scan}
T_n=\max_{\substack{u\in\mathcal U_n\\\widehat F_n^p(u)>u}}
\left\{n\operatorname{kl}\bigl(\widehat F_n^p(u),u\bigr)
                 +\log w_n(u)\right\},\ 
\phi_{n,\alpha}=\ind\{T_n>\log(1/\alpha)\}.
\end{equation}
Each term in \eqref{eq:scan} is the log-likelihood ratio statistic for an excess of small $p$-values at level $u$, penalized by its weight. The restriction to $\widehat F_n^p(u)>u$ reflects the fact that watermarking can only inflate the lower tail of the $p$-values. The resulting statistic is a weighted, one-sided version of the Berk--Jones statistic \citep[Section 2]{jager2007}, and its scan over several tail levels follows the higher-criticism principle of \citet{donoho2004}.

With these ingredients in place, we can now describe the complete detection procedure, summarized in Algorithm~\ref{alg:test}. Given a text and the secret key, the detector first reconstructs, at each retained position, the partition, the Gumbel table, and the selected column, exactly as is required for $T_n^\tg$. From these, it computes the pivots $Y_t$ and the $p$-values $p_t=1-Y_t$. It then evaluates the empirical tail proportions $\widehat F_n^p(u)$ on the grid $\mathcal U_n$, forms $T_n$ as in \eqref{eq:scan}, and declares the text watermarked when $T_n$ exceeds $\log(1/\alpha)$. The procedure therefore uses the same inputs as the report's detector, and in particular requires no knowledge of the NTPs $P_t$, the entropy fractions $\beta_t$, the edit fraction $\varepsilon_n$, or the transport coupling.

\begin{algorithm}[h!]
\caption{Adaptive detection of \tg\ by a weighted binomial scan}
\label{alg:test}
\KwInput{Text, tokenizer, secret key, shared partition and column
configuration, context rule, and significance level $\alpha\in(0,1)$.}
\KwOutput{Detection or no detection.}
Tokenize the text and retain the first scored occurrence of each context
window prescribed by the shared context rule\;
Let $n$ be the number of retained positions\;
\If{$n=0$}{\Return no detection\;}
\For{$t=1,\ldots,n$}{
Reconstruct the partition $g_t$, cost table $(Z_{t,bj})_{b,j}$, and column $J_t$
from the key and observed context\;
Set $Y_t=F_G(Z_{t,g_t(W_t),J_t})$ and $p_t=1-Y_t$ as in
\eqref{eq:score_convention}\;
}
Construct $\mathcal U_n$ and $w_n$ from
\eqref{eq:grid}--\eqref{eq:weights}\;
Sort $p_1,\ldots,p_n$ and compute
$\widehat F_n^p(u)=n^{-1}\#\{t:p_t\le u\}$ for each $u\in\mathcal U_n$\;
Compute $T_n$ from \eqref{eq:scan}\;
\eIf{$T_n>\log(1/\alpha)$}{\Return detection\;}{\Return no detection\;}
\end{algorithm}

After score reconstruction, the procedure requires only a sort of the $p$-values, and hence $O(n\log n)$ operations. The next theorem establishes its finite-sample validity and a general sufficient condition for power.

\begin{theorem}[Size and conditional power]
\label{thm:test}
Let $n\ge1$ be a deterministic integer, and suppose that the conditional
score laws satisfy \eqref{eq:mixture}, with a deterministic
$\varepsilon_n\in(0,1]$, with respect to a filtration to
which the scores are adapted. For every $\alpha\in(0,1)$, the test
$\phi_{n,\alpha}$ in Algorithm~\ref{alg:test}, defined by
\eqref{eq:grid}, \eqref{eq:weights}, and \eqref{eq:scan}, satisfies
$\Pp_0(\phi_{n,\alpha}=1)\le\alpha$.
For the power statement, suppose the conditional signal densities
satisfy $0\le f_{1,t}\le M$ almost surely, where $1\le M<\infty$ is
deterministic and may depend on $n$. Suppose also that there are
deterministic $u\in\mathcal U_n$ and $d>0$ such that
$\int_{1-u}^1f_{1,t}(y)\dd y\ge u+d$ almost surely for every
$1\le t\le n$. If
\[
 \frac{n\varepsilon_n^2d^2}{8Mu}
 >\log\frac1{\alpha w_n(u)},
\]
then its type II error satisfies
\[
 \Pp_1(\phi_{n,\alpha}=0)
 \le\min\left\{1,\frac{4Mu}{n\varepsilon_n^2d^2}\right\},
\]
with both bounds being uniform over all conditional laws satisfying the
stated assumptions. 
\end{theorem}

In Theorem \ref{thm:test}, the size guarantee is exact at every sample size and holds under arbitrary dependence across positions, since the weights act as a weighted Bonferroni correction over the grid. Power reduces to finding a single level $u\in\mathcal U_n$ at which the signal places excess tail mass $d$. The factor $u$ in the condition favours signals concentrated in small tails, which are exactly those that $T_n^\tg$ dilutes, and the price of not knowing $u$ is only $\log(1/w_n(u))=O(\log\log n)$. The remaining results verify this condition in each regime, showing that the test recovers every detection boundary of Section~\ref{sec:model},

We begin with the fixed-column regime of Theorem~\ref{thm:boundary}, where the report's sum is already optimal. Applying Theorem~\ref{cor:tokens} to the nondecreasing score $h(y)=\ind\{y>1/2\}$ shows that the signal places excess mass of the order of the square root of the information above the median. The weight reserved for $u=1/2$ in \eqref{eq:weights} therefore suffices, and the following theorem shows that the proposed test attains the boundary of Theorem~\ref{thm:boundary} without knowledge of $\varepsilon_n$, $\Delta_n$, or $\beta_n$.

\begin{theorem}[Adaptive optimality at the detection boundary]
\label{thm:boundary_adaptive}
Consider the token-class setup of Theorem~\ref{thm:boundary}.
For every fixed $\alpha\in(0,1)$,
Algorithm~\ref{alg:test} satisfies $\Pp_0(\phi_{n,\alpha}=1)\le\alpha$
for every $n$ and
\[
 \sup_{\Pp_1\in\cH_{1,n}(\Delta_n,\beta_n)}
       \Pp_1(\phi_{n,\alpha}=0)\to0
\]
whenever $n\varepsilon_n^2\Delta_n
\min\{\Delta_n,\beta_n\mathbb H(\Delta_n)\}\to\infty$.
Fix $\alpha_0\in(0,1)$ and let $\alpha_n=\alpha_0/\log(n+e)$.
If
\[
 \frac{n\varepsilon_n^2\Delta_n
       \min\{\Delta_n,\beta_n\mathbb H(\Delta_n)\}}
      {\log\log(n+e)}\to\infty,
\]
then its sum of type I and worst-case type II errors satisfies
\[
 \Pp_0(\phi_{n,\alpha_n}=1)
 +\sup_{\Pp_1\in\cH_{1,n}(\Delta_n,\beta_n)}
       \Pp_1(\phi_{n,\alpha_n}=0)\to0.
\]
\end{theorem}

Theorem~\ref{thm:boundary_adaptive} shows that adaptivity comes for free at the fixed-column boundary: the fixed-level test is consistent under exactly the necessary and sufficient condition of Theorem~\ref{thm:boundary}, because the median weight keeps the penalty $\log\{1/(\alpha w_n(1/2))\}$ bounded. Driving the type I error to zero as well costs only a $\log\log n$ factor, as for the report's sum in Theorem~\ref{thm:sums}. Hence, adopting the proposed test in place of $T_n^{\tg}$ entails no loss in the regime where the report's statistic is optimal. Specializing to the polynomial scalings of Corollary~\ref{cor:phase}, a single vanishing-level test attains the entire consistency region of the phase diagram.

\begin{corollary}[Adaptive attainment of the phase diagram]
\label{cor:phase_adaptive}
Under the token-class assumptions of Corollary~\ref{cor:phase}, fix
$\alpha_0\in(0,1)$ and let $\alpha_n=\alpha_0/\log(n+e)$.
If $2p+2q+r<1$, or if $2p+2q+r=1$ with $q>0$ and $r>0$, then
Algorithm~\ref{alg:test} has vanishing sum of type I and worst-case
type II errors:
\[
 \Pp_0(\phi_{n,\alpha_n}=1)
 +\sup_{\Pp_1\in\cH_{1,n}(\Delta_n,\beta_n)}
       \Pp_1(\phi_{n,\alpha_n}=0)\to0.
\]
\end{corollary}

The region in Corollary~\ref{cor:phase_adaptive} coincides with that of Corollary~\ref{cor:phase}. It includes the part of the boundary surface $2p+2q+r=1$ with $q,r>0$, where the information diverges only logarithmically, and on the rest of that surface no test is consistent. We now turn to the growing-column regimes, where the report's sum is no longer optimal. When the number of columns grows, the entropy budget determines where the signal lies. A large budget allows it to concentrate in a small upper tail, while a weak budget keeps it visible at the median. Since the grid contains both the dyadic levels and the median, a single test covers both cases.

\begin{theorem}[Adaptive attainment with growing columns]
For a fixed $\alpha_0\in(0,1)$ and $\alpha_n=\alpha_0/\log(n+e)$, the test of Algorithm~\ref{alg:test} satisfies
\[
 \Pp_0(\phi_{n,\alpha_n}=1)
 +\sup_{\Pp_1\in\mathcal H}
       \Pp_1(\phi_{n,\alpha_n}=0)\to0
\]
in each of the following cases.
\begin{enumerate}
\item[(a)] \label{thm:activehigh_adaptive} \textup{(Large budget.)} $\mathcal H=\cH_{1,n}^{\mathrm{high}}$, under the setup of Theorem~\ref{thm:activehigh} with $\varepsilon_n\asymp n^{-p}$, $p\ge0$, and $2p+2q-\min(s,q)<1$.
\item[(b)] \label{thm:activelow_adaptive} \textup{(Weak budget.)} $\mathcal H=\cH_{1,n}^{\mathrm{low}}$, under the setup of Theorem~\ref{thm:activelow} with $n\varepsilon_n^2\Delta_n\kappa_n/\log\log(n+e)\to\infty$. 
\end{enumerate}
\end{theorem}

Theorem~\ref{thm:activehigh_adaptive} completes the answer to our second question. Part (a) attains the full detectable region of Theorem~\ref{thm:activehigh}, including the region $\{p+q>1/2,\ 2p+2q-\min(s,q)<1\}$ where, by Proposition~\ref{prop:sumloss}, every threshold for $T_n^{\tg}$ fails. Part (b) matches the weak-budget boundary of Theorem~\ref{thm:activelow}, as the report's sum does by Theorem~\ref{thm:activelow_sum}. The proposed test therefore remains optimal in every regime considered, including those where $T_n^{\tg}$ is not. 

The tests $\psi_n^*$ in Theorem~\ref{thm:efficiency} are built from the least-favourable density in $\mathcal Q_\Theta$, and therefore require knowledge of $\Theta$, which is unavailable in practice. The proposed test avoids this requirement. Since the edited score laws are nondecreasing, their tail probabilities on the increasingly fine mesh in \eqref{eq:grid} determine them asymptotically, and the next theorem shows that this suffices to attain the optimal exponent.

\begin{theorem}[Adaptive attainment of the optimal exponential efficiency]
\label{thm:efficiency_adaptive}
Under the setup of Theorem~\ref{thm:efficiency}, for every $\alpha_n\in(0,1)$ satisfying $\limsup_n\alpha_n<1$ and $\log(1/\alpha_n)=o(n)$, the test $\phi_{n,\alpha_n}$ of Algorithm~\ref{alg:test} satisfies
\[
 \lim_{n\to\infty}-\frac1n\log
 \sup_{\Pp_1\in\cH_{1,n}(\Theta)}
       \Pp_1(\phi_{n,\alpha_n}=0)
 =\inf_{q\in\mathcal Q_\Theta}
       \left\{-\int_0^1\log q(y)\dd y\right\}.
\]
Consequently, under the positivity conditions of Theorem~\ref{thm:efficiency} and $\alpha_n\to0, \Pp_0(\phi_{n,\alpha_n}=1)
 +\sup_{\Pp_1\in\cH_{1,n}(\Theta)}
       \Pp_1(\phi_{n,\alpha_n}=0)\to0.$
\end{theorem}

Theorem~\ref{thm:efficiency_adaptive} settles our third question: the proposed test attains the optimal exponent of Theorem~\ref{thm:efficiency} uniformly over $\cH_{1,n}(\Theta)$, without knowledge of $\Theta$ or of the least-favourable density. In particular, at every fixed \tg\ alternative of Proposition~\ref{prop:effsum}, it attains the exponent that the report's sum misses. Combined with Theorems~\ref{thm:boundary_adaptive} and~\ref{thm:activehigh_adaptive}, a single parameter-free test is therefore optimal both at the detection boundary and in exponential efficiency, in every regime considered.

\section{Discussion}
\label{sec:discussion}
This paper set out to understand what the entropy calibration of \tg\ costs at detection time, and whether the sum statistic proposed by \citet{textgrain2026} remains an adequate detector once the text has been edited. We found that \tg\ is strictly harder to detect than Gumbel-max and maximal-coupling watermarks (Corollary~\ref{cor:phase}), a cost that stems from its construction rather than from preserving entropy itself (Example~\ref{ex:matched-entropy-separation}), and that this very loss makes the sum statistic optimal at the detection boundary when the number of columns is fixed (Theorem~\ref{thm:sums}). However, this optimality is not the whole picture. Indeed, the sum statistic misses detectable alternatives when the number of columns grows under a large entropy budget (Proposition~\ref{prop:sumloss}), and it is never exponentially efficient at a fixed edit fraction (Proposition~\ref{prop:effsum}). The weighted multiscale binomial scan of Algorithm~\ref{alg:test} closes both gaps, attaining every detection boundary and the optimal exponential rate without knowledge of the next-token distributions, the entropy budget, or the edit fraction (Theorems~\ref{thm:boundary_adaptive}, \ref{thm:activehigh_adaptive} and~\ref{thm:efficiency_adaptive}), although extending these guarantees to the repeated contexts of the official implementation in \cite{textgrain2026} remains open.

\section*{AI usage statement\footnote{\href{https://leidendeclaration.ai/}{https://leidendeclaration.ai/}}}
The authors conceived the ideas and the research questions, and formulated the results. They acknowledge the use of ChatGPT-Pro 6.1 Astra and Claude Opus 5.5 for sharpening theorem statements and general exposition. The authors retain full responsibility for the veracity of the content.

\small
\bibliographystyle{plainnat}
\bibliography{references}

\normalsize

\clearpage
\appendix
\begin{center}
    \Large Appendix
\end{center}
This appendix contains a broad overview of literature related to our works, along with an auxiliary result validating Example \ref{ex:matched-entropy-separation}. Detailed proofs of our results are available upon request, and will be included in the forthcoming extended version. 
\section{Related Literature}
Our work sits at the confluence of two strands of research: the design of watermarking schemes for LLMs, and the statistical theory of their detection.

On the design front, there are two desiderata: fidelity to the language model, and stochasticity of generation. . For example, Red-Green watermarking \citep{kirchenbauer2023watermark}, since adapted well beyond text \citep[e.g.,][]{zhou2026green}, injects additional stochasticity onto the sampled tokens under watermarking, which reduces repetition. While effective, implementing this method distorts the next token probability distribution (NTP), from which the tokens are sampled; in other words, Red-Green watermark is \textit{biased}. On the other hand, Gumbel-max watermarking \citep{aaronson2023watermarking} is \textit{unbiased} (i.e. a watermarked token has, marginally, the same NTP as an unwatermarked one), but remains deterministic given the pseudorandom variable implementing the watermark. Subsequent schemes preserve unbiasedness through a variety of constructions, including inverse-transform sampling \citep{kuditipudi2024robust}, distribution-preserving reweighting \citep{hu2024unbiased, wu2024dipmark}, tournament sampling \citep{dathathri2024scalable}, cryptographically undetectable sampling \citep{christ2024undetectable}, and permute-and-flip decoding \citep{zhao2025permute}. Closest to \tg\ is the scheme of \citet{xie2025debiasing}, which removes the bias of Red-Green watermarking through a maximal coupling while retaining its block structure. \tg\ \citep{textgrain2026} replaces this coupling by the entropic OT coupling \eqref{eq:transport}, in which the budget $\beta$ caps the dependence between the keyed table and the sampled block, measured by $\KL(\pi\|P^{\blk}\otimes m^{-1}\bo)$, as a fraction of $H(P)$.

On the detection front, \citet{li2025framework} cast watermark detection as a hypothesis test on pivotal statistics, and introduced a class-dependent efficiency criterion that yields optimal score functions for Gumbel-max. \citet{li2026robust} model human edits through the mixture alternative underlying \eqref{eq:mixture}, identify the Gumbel-max detection boundary $2p+q=1$, show that sum-based tests are strictly suboptimal on it, and propose an adaptive truncated goodness-of-fit test that attains it. \citet{xie2025debiasing} carry out the analogous program for maximal coupling, where the boundary is again $2p+q=1$, whereas the sum test only reaches $p+q=1/2$. Related work addresses pseudorandom collisions under repeated contexts \citep{cai2025optimal}, localization of watermarked segments in mixed-source text \citep{blanchet2026optimal}, estimation of the watermarked proportion \citep{li2025optimal}, and the detection--distortion trade-off \citep{cai2026towards}. We follow the template of \citet{li2025framework, li2026robust}; however, the \tg\ alternative carries two parameters absent from these works, namely the entropy budget $\beta$ and the number of columns $m$. Both reshape the boundary itself (Corollary~\ref{cor:phase} and Theorem~\ref{thm:activehigh}), and with $m$ fixed they reverse the verdict on the sum test (Theorem~\ref{thm:sums}) compared to previous literature.

Finally, the sparse regime $\varepsilon_n\to0$ ties our problem to the detection of sparse heterogeneous mixtures \citep{donoho2004, donoho2015}. There, Fisher's combination test, of which $2T_n^{\tg}$ is an instance, fails in the very sparse regime, while higher criticism and $\phi$-divergence statistics attain the boundary adaptively \citep{jager2007, ditzhaus2019}. The multiscale test of \S\ref{subsec:test} is a weighted, one-sided Berk--Jones statistic in this tradition. For a fixed edit fraction, the efficiency analysis of \S\ref{sec:efficiency} becomes a Chernoff--Stein problem against a convexified composite alternative, in the spirit of the reverse information projection of \citet{larsson2025}.

\section{Preserving entropy need not lead to harder detection}
 The following proposition gives the precise comparison used in Example \ref{ex:matched-entropy-separation}. 
\begin{proposition}
\label{prop:example}
Consider Example~\ref{ex:matched-entropy-separation}. Recall that $a_{\delta}$ satisfies
\begin{align}
    c\delta = \mathbb{H}(\delta)
 -(1-\delta)\mathbb{H}((1-a)\delta) -\delta \mathbb{H}(a+(1-a)\delta). \label{eq:a-del}
\end{align}
Write $Q^{\tg}_\delta$ for the \tg\ watermarking employing the same NTP, regularization and partition,
$(P_\delta,\beta_\delta,g)$. For $\Delta_n=n^{-2/3}$ and $\varepsilon_n=1$, let $\{ F_G(Z_{t,g(W_t),J_t})\}_{t=1}^n \sim \IP_{k,n}$  under
decoder $k$, and
$\Pp_0=\Unif(0,1)^{\otimes n}$. Then
\[
\begin{aligned}
 \inf_{\phi_n}\{\E_0\phi_n+\E_{\tg,n}(1-\phi_n)\}
 \to1,\ \text{and }
 \inf_{\phi_n}\{\E_0\phi_n+\E_{\mathrm{RG},n}(1-\phi_n)\}
 \to0.
\end{aligned}
\]
\end{proposition}

\end{document}